\def\year{2019}\relax
\documentclass[letterpaper]{article} 
\usepackage{aaai19}  
\usepackage{times}  
\usepackage{helvet}  
\usepackage{courier}  
\usepackage{url}  
\usepackage{graphicx}  
\usepackage{subcaption}
\usepackage{amsmath}
\usepackage{color,soul}
\graphicspath{{./figures/}} 
\usepackage{algorithm} 
\usepackage[noend]{algpseudocode} 
\renewcommand{\vec}[1]{\textbf{\textit{#1}}} 
\usepackage{soul}
\usepackage{flushend}
\usepackage{enumerate}

\DeclareMathOperator*{\argmin}{argmin}

\definecolor{gray}{RGB}{200,200,200}
\definecolor{orange}{RGB}{237,125,49}

\begin{document}
%





\title{Efficiently Combining Human Demonstrations and Interventions for Safe Training of Autonomous Systems in Real-Time}

\author{Vinicius G. Goecks\textsuperscript{1,2}, Gregory M. Gremillion\textsuperscript{1}, Vernon J. Lawhern\textsuperscript{1}
\AND
John Valasek\textsuperscript{2}, Nicholas R. Waytowich\textsuperscript{1,3}\\
\textsuperscript{1}{US Army Research Laboratory},
\textsuperscript{2}{Texas A\&M University},
\textsuperscript{3}{Columbia University}\\
\texttt{vinicius.goecks@tamu.edu},\\
\texttt{\{gregory.m.gremillion, vernon.j.lawhern\}.civ@mail.mil}\\
\texttt{valasek@tamu.edu, nicholas.r.waytowich.civ@mail.mil}\\
}

\maketitle 
\begin{abstract}
This paper investigates how to utilize different forms of human interaction to safely train autonomous systems in real-time by learning from both human demonstrations and interventions. 
We implement two components of the Cycle-of-Learning for Autonomous Systems, which is our framework for combining multiple modalities of human interaction.
The current effort employs human demonstrations to teach a desired behavior via imitation learning, then leverages intervention data to correct for undesired behaviors produced by the imitation learner to teach novel tasks to an autonomous agent safely, after only minutes of training. 
We demonstrate this method in an autonomous perching task using a quadrotor with continuous roll, pitch, yaw, and throttle commands and imagery captured from a downward-facing camera in a high-fidelity simulated environment. 
Our method improves task completion performance for the same amount of human interaction when compared to learning from demonstrations alone, while also requiring on average 32\% less data to achieve that performance.
This provides evidence that combining multiple modes of human interaction can increase both the training speed and overall performance of policies for autonomous systems. 
\end{abstract}

\section{Introduction}
\noindent The primary goal of learning methodologies is to imbue intelligent agents with the capability to autonomously and successfully perform complex tasks, when \emph{a priori} design of the necessary behaviors is intractable. 
Most tasks of interest, especially those with real-world applicability, quickly exceed the capability of designers to handcraft optimal or even successful policies.
It can even be infeasible to construct appropriate objective or reward functions in many cases.
Instead, learning techniques can be used to empirically discover the underlying objective function for the task and the policy required to satisfy it, typically utilizing state, action, or reward data.
Several classes of these techniques have yielded promising results, including learning from demonstration, learning from evaluation, and reinforcement learning.


Reinforcement learning has been proven to work on scenarios with well-designed reward functions and easily available interactions with the environment \cite{Mnih2015a}. 
However, in real-world robotic applications, explicit reward functions are non-existent, and interactions with the hardware are expensive and susceptible to catastrophic failures. 
This motivates leveraging human interaction to supply this reward function and task knowledge, to reduce the amount of high-risk interactions with the environment, and to safely shape the behavior of robotic agents.

Learning from evaluation is one such way to leverage human domain knowledge and intent to shape agent behavior through sparse interactions in the form of evaluative feedback, possibly allowing for the approximation of a reward function \cite{Knox2009,MacGlashan2017,Warnell2018}.
This technique has the advantage of minimally tasking the human evaluator and can be used when training behaviors they themselves cannot perform.
However, it can be slow to converge as the agent can only identify desired or even stable behaviors through more random exploration or indirect guidance from human negative reinforcement of unwanted actions, rather than through more explicit examples of desired behaviors.

In such a case, learning from demonstration can be used to provide a more directed path to these intended behaviors by utilizing examples of the humans performing the task.
This technique has the advantage of quickly converging to more stable behaviors.
However, given that it is typically performed offline, it does not provide a mechanism for corrective or preventative inputs when the learned behavior results in undesirable or catastrophic outcomes, potentially due to unseen states.
Learning from demonstration also inherently requires the maximal burden on the human, requiring them to perform the task many times until the state space has been sufficiently explored, so as to generate a robust policy. 
Also, it necessarily fails when the human is incapable of performing the task successfully at all.

Learning from interventions, where a human acts as an overseer while an agent is performing a task and periodically takes over control or intervenes when necessary, can provide a method to improve the agent policy while preventing or mitigating catastrophic behaviors \cite{Saunders2017}.
This technique can also reduce the amount of direct interactions with the agent, when compared to learning from demonstration.
Similar to learning from evaluation, this technique suffers from the disadvantage that desired behaviors must be discovered through more variable exploration, resulting in slower convergence and less stable behavior.

Most of these human interaction methods have been studied separately, and there is very little work combining multiple modalities to leverage strengths and mitigate weaknesses. 
In this paper, we work towards our conceptual framework that combines multiple human-agent interaction modalities into a single framework, called the Cycle-of-Learning for Autonomous Systems from Human Interaction \cite{Waytowich2018}. 
Our goal is to unify different human-in-the-loop learning techniques in a single framework to overcome the drawbacks of training from different human interaction modalities in isolation, while also maintaining data-efficiency and safety. 

In this paper, we present our initial work towards this goal with a method for combining learning from demonstrations and learning from interventions for safe and efficient training of autonomous systems. We seek to develop a real-time learning technique that combines demonstrations as well as interventions provided from a human to outperform traditional imitation learning techniques while maintaining agent safety and requiring less data. We validate our method with an aerial robotic perching task in a high-fidelity simulator using a quadrotor that has continuous roll, pitch, yaw and throttle commands and a downward facing camera. In particular, the contributions of our work are twofold:
\begin{enumerate}[\indent (1)]
	\item  We propose a method for efficiently and safely learning from human demonstrations and interventions in real-time. 
	\item We empirically investigate both the task performance and data efficiency associated with combining human demonstrations and interventions.
\end{enumerate}
We show that policies trained with human demonstrations and human interventions together outperform policies trained with just human demonstrations while simultaneously using less data. To the best of our knowledge this is the first result showing that training a policy with a specific sequence of human interactions (demonstrations, then interventions) outperforms training a policy with just human demonstrations (controlling for the total amount of human interactions), and that one can obtain this performance with significantly reduced data requirements, providing initial evidence that the role of the human should adapt during the training of safe autonomous systems.

\begin{figure}[!t]
    \centering
    \includegraphics[width=1.0\linewidth]{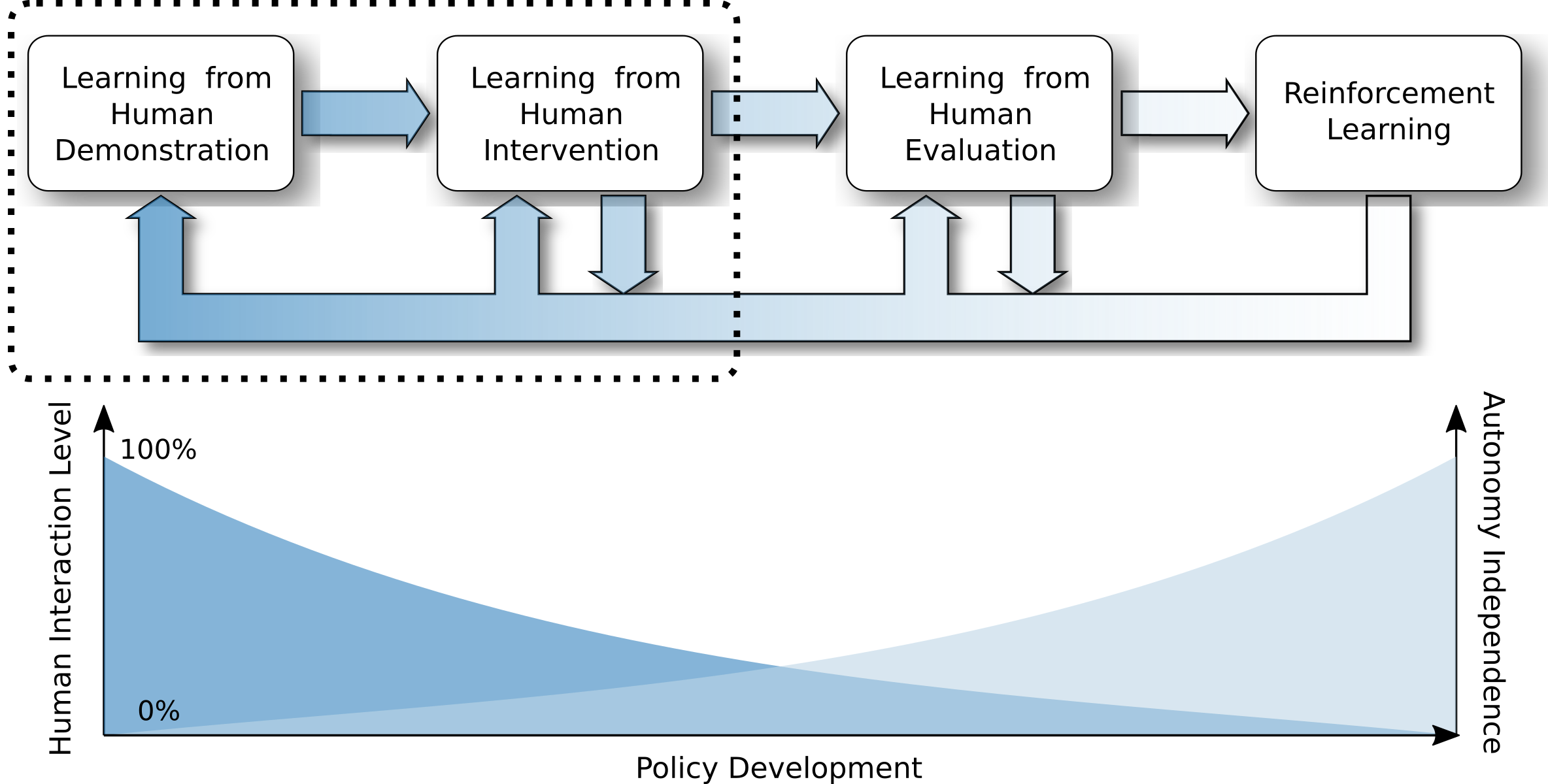}
    \caption{Cycle-of-Learning for Autonomous Systems from Human Interaction: a concept for combining multiple forms of human interaction with reinforcement learning. As the policy develops, the autonomy independence increases and the human interaction level decreases. This work focuses on the first two components of the cycle (dashed box): Learning from Demonstration and Learning from Intervention.}
    \label{fig:col_diagram}
\end{figure}

\section{Background and Related Work}
\subsection{Learning from Demonstrations}
Here we provide a brief summary of Learning from Demonstrations (LfD); a more comprehensive review can be found in \cite{Argall2009}. 
Learning from Demonstrations, sometimes referred to as \emph{Imitation Learning}, is defined by training a policy $\pi$ in order to generalize over a subset $\mathcal{D}$ of states and actions visited during a task demonstration over $T$ time steps:
\begin{align*} \label{eq:demonstration_dataset}
\mathcal{D} = \left\{\vec{a}_0, \vec{s}_0, \vec{a}_1, \vec{s}_1, ... , \vec{a}_T, \vec{s}_T\right\}.
\end{align*}
This demonstration can be performed by a human supervisor, optimal controller, or virtually any other pre-trained policy.

In the case of human demonstrations, the human is implicitly trying to maximize what may be represented as an internal reward function for a given task (Equation \ref{eq:imitation_rew}), where $\pi^*(\vec{a}^*_t | \vec{s}_t)$ represents the optimal policy that is not necessarily known, in which the optimal action $\vec{a}^*$ is taken at state $\vec{s}$ for every time step $t$.

\begin{equation}\label{eq:imitation_rew}
\max_{\vec{a}_0,...,\vec{a}_T} \sum_{t=0}^{T} r_t(\vec{s}_t, \vec{a}_t) = \sum_{t=0}^{T} \log p (\pi^*(\vec{a}_t^* | \vec{s}_t))
\end{equation}

Defining the policy of the supervisor as $\pi_{sup}$ and its estimate as $\hat{\pi}_{sup}$, imitation learning can be achieved through standard supervised learning, where the parameters $\theta$ of a policy $\pi_\theta$ are trained in order to minimize a loss function, such as mean squared error, as shown in Equation \ref{eq:sup_mse}.

\begin{equation}\label{eq:sup_mse}
\hat{\pi}_{sup} = \argmin_{\pi_\theta} \sum_{t=0}^{T} ||\pi_\theta(\vec{s}_t) - \vec{a}_t ||^2
\end{equation}

There are many empirical successes of using imitation learning to train autonomous systems. For self-driving cars, \citeauthor{Bojarski2016} successfully used human demonstrations to train a policy that mapped from front-facing camera images to steering wheel commands using around one hundred hours of human driving data \cite{Bojarski2016}. 
Similar approaches have been taken to train small unmanned air system (sUAS) to navigate through cluttered environments while avoiding obstacles, where demonstration data was collected by human oracles in simulated \cite{Goecks2018} and real-world environments \cite{Giusti2015}.

\subsection{Learning from Interventions}
In \emph{Learning from Interventions} (LfI) the human takes the role of a supervisor and watches the agent performing the task and intervenes (i.e. overriding agent actions with human actions) when necessary, in order to avoid unsafe behaviors that may lead to catastrophic states. 
Recently, this learning from human intervention concept was used for safe reinforcement learning (RL) that could train model-free RL agents without a single catastrophe \cite{Saunders2017}. Similar work has proposed using human interaction to train a classifier to detect unsafe states, which would then trigger the intervention by a safe policy previously trained based on human demonstration of the task \cite{Hilleli2018}. This off-policy data generated by the safe policy is aggregated to the replay buffer of a value-based reinforcement learning algorithm (Double Deep Q-Network, or DDQN \cite{Hasselt2015}). The main advantage of this method is being able to combine the off-policy data generated by the interventions to update the current policy.

\subsection{Related Work}

Several existing works have studied, in isolation, the use of different human interaction modalities to train policies for autonomous systems, either in the form of demonstrations \cite{Akgun2012}, \cite{Argall2009}, interventions \cite{Akgun2012a}, \cite{Saunders2017} or evaluations \cite{Knox2009}.  However, there has been relatively little work on how to effectively combine multiple human interaction modalities into a single learning framework. Several cases include the combination of demonstrations and mixed initiative control for training robot polices \cite{Grollman2007} as well as the recent work by \citeauthor{Hilleli2018} where imitation learning was combined with interactive reward shaping in a simulated racing game  \cite{Hilleli2018} and the recent work \cite{Peng2018} where deviation from the expert demonstration is added to a reward function to be optimized with reinforcement learning.

Another example of work that attempts to augment learning from demonstrations with additional human interaction is the Dataset Aggregation (DAgger) algorithm \cite{Ross2011}.
DAgger is an iterative algorithm that consists of two policies, a primary agent policy that is used for direct control of a system, and a reference policy that is used to generate additional labels to fine-tune the primary policy towards optimal behavior. Importantly, the reference policy's actions are not taken, but are instead aggregated and used as additional labels to re-train the primary policy for the next iteration. In \cite{Ross2013} DAgger was used to train a collision avoidance policy for an autonomous quadrotor using imitation learning on a set of human demonstrations to learn the primary policy and using the human observer as a reference policy. 
There are some drawbacks to this approach that are worth discussing. As noted in \cite{Ross2013}, because the human observer is never in direct control of the policy, safety is not guaranteed, since the agent has the potential to visit previously unseen states, which could cause catastrophic failures. 
Additionally, the subsequent labeling by the human can be suboptimal both in the amount of data recorded (perhaps recording more data in suboptimal states than is needed to learn an optimal policy) as well as in capturing the intended result of the human observer's action (as in distinguishing a minor course correction from a sharp turn, or the appropriate combination of actions to perform a behavior). 
Another limitation of DAgger is that the human feedback was provided \emph{offline} after each run while viewing a slower replay of the video stream to improve the resulting label quality. 
This prevents the application to tasks where real-time interaction between humans and agents are required.

\section{Proposed Methodology: Cycle-of-Learning}
This work demonstrates a technique for efficiently training an autonomous system safely and in real-time by combining learning from demonstrations and interventions. It is the first part of the Cycle-of-Learning concept (Figure~\ref{fig:col_diagram}) which aims to combine multiple forms of human-agent interaction for learning a policy that mimics the human trainer in a safe and efficient manner. Although this paper focuses on the first two parts of the Cycle-of-Learning, for brevity, we will refer to the algorithm presented here as the Cycle-of-Learning (CoL). 

The CoL starts by training an initial policy $\pi_0$ from a set of task demonstrations provided by the human trainer using a standard supervised learning technique (regression in this case since the action-space for our task is continuous). 
Next, the agent is given control and executes $\pi_0$ while the human takes the role of overseer and supervises the agent's actions. 
Using a joystick controller, the human intervenes whenever the agent exhibits unwanted behavior that diverges from the policy of the human trainer, and provides corrective actions to drive the agent back on course, and then releases control back to the agent. 
The agent then learns from this intervention by augmenting the original training dataset with the states and actions from the intervention, and then fine-tuning $\pi_0$. 
The agent then executes the new policy $\pi_n$ while the human continues to oversee and provides interventions as necessary. 
In practice, the human trainer can easily switch between providing demonstrations and interventions by switching control between the human and the agent as shown in Figure \ref{fig:airsim_col_diagram_compact}.
Combining demonstration and intervention data in this way should not only improve the policy over what learning from demonstration can do alone but also require less training data to do so. 
The intuition is that the agent will inevitably end up in states previously unexplored with the original demonstration data which will cause it's policy to fail and that intervening from those failure states allows the agent to quickly learn from those deficiencies or "blind spots" in its own policy in a more targeted fashion than from demonstration data alone \cite{Ramakrishann2018}. 
In this way, we learn only from the critical states, which is more data efficient, instead of using all states for training as is done in DAgger \cite{Ross2011}.

\subsection{Data Efficiency}
A demonstration is defined as a human-produced trajectory of state-action pairs for the entire episode, while an intervention is defined as a trajectory of state-action pairs for only the subset of the episode where corrective action is deemed necessary by the human. 
Thus, the amount of data provided via intervention is nearly always less than the amount provided via demonstration. 
Training routines that incorporate more episodes utilizing learning from intervention rather than learning from demonstration will in general be more data sparse, assuming comparable task performance.
Therefore, by utilizing components of the CoL to learn from both demonstration and intervention, we can train with less data than if demonstrations had been used in isolation for an equivalent number of episodes, resulting in a more efficient training framework.
This concept generalizes to the full CoL, as the agent naturally requires less input from the human as its policy develops, its task proficiency increases, and it becomes more autonomous (indicated in Figure \ref{fig:col_diagram}).

\subsection{Safe Learning}
The notion of \emph{safe} learning here refers to the ability of a human oracle to intervene in cases where catastrophic failure may be imminent.
Thus, the agent is able to explore higher risk regions of the state space with a greater degree of safety.
This approach leverages human domain knowledge and ability to forecast such boundary states, which the agent cannot do early in the training process when the state space is less explored.
By allowing the policy to explore less seen regions and then provide training data of how to correct from those states, human interventions provide a richer dataset that improves the policy in those regimes.
This is contrasted to a method based solely on demonstration, which may only see states and observations along a nominal trajectory and have a policy poorly fit to data outside that envelope.
The result is a policy that is more robust, through greater data diversity, while not risking damage to the agent that is typical with methods that rely on random exploration of the state space.
This provides a method to safely train an autonomous system.

\subsection{Real-Time Interaction}
The utility of the demonstrated approach is partially linked to the ability of the agent to consume data as it is provided by the human oracle and update its policy online.
The current system accomplishes this by storing all subject state-action pairs in the training dataset, which is queried in real-time to update the policy, and then fine-tuning that policy whenever new samples are added to the dataset.
During intervention, this allows for interaction with an agent using a policy trained on the most recent corrective actions provided by the human.
The short time between novel human intervention data and behavioral roll-outs from the agent policy prevents significant delay in this feedback loop that might result from more infrequent, batch learning.
As in closed loop systems, large temporal delays between feedback inputs and their resultant output behaviors can lead to instability.
In this context, that would manifest as unstable training as the human oracle would need to correct for undesired actions for significantly longer before seeing any effect on the agent behavior.
This shortcoming was exhibited in DAgger, where policy correction was a delayed, offline process.

\begin{figure}[!t]
    \centering
    \includegraphics[width=1.0\linewidth]{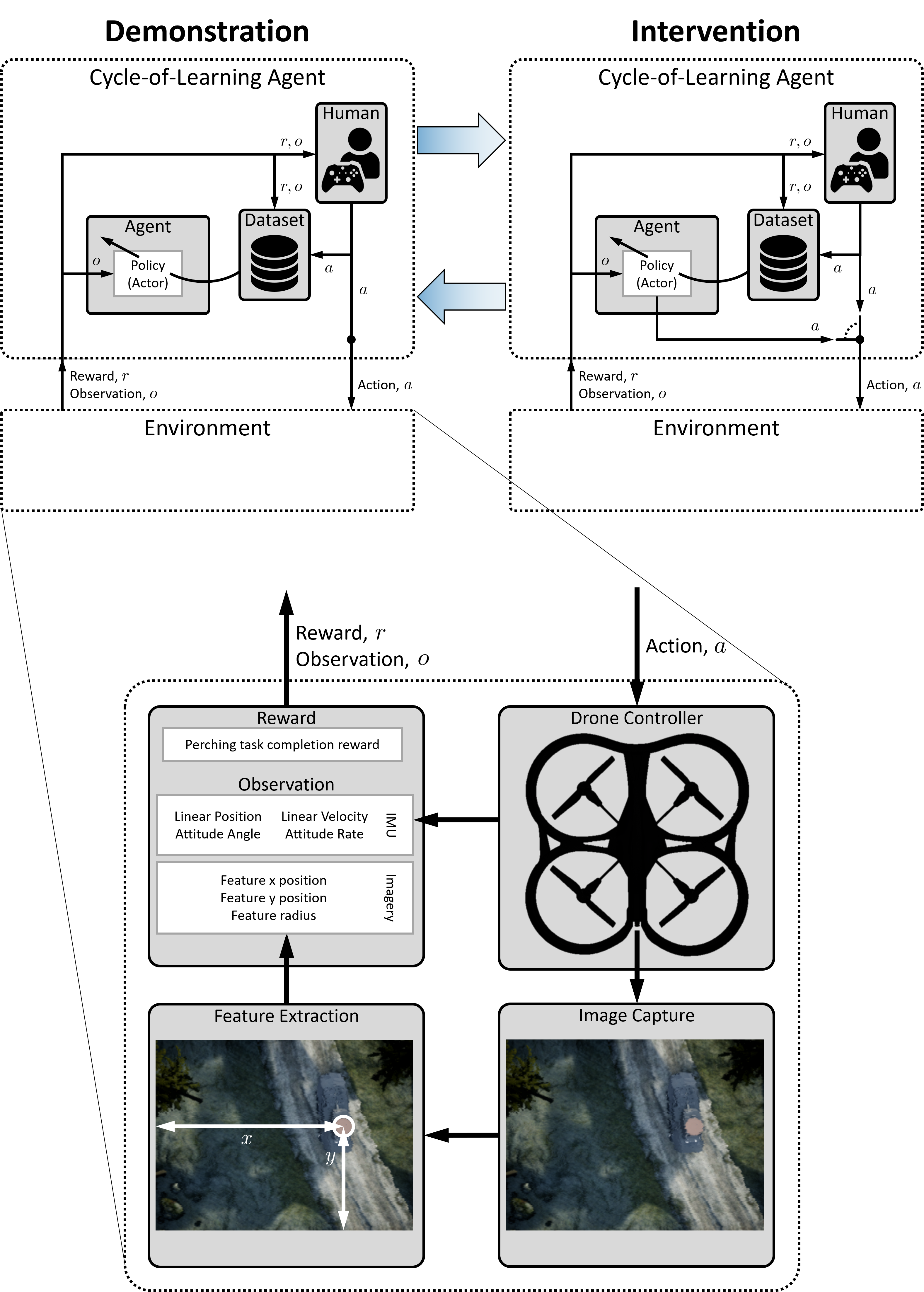}
    \caption{Flow diagram illustrating the learning from demonstration and intervention stages in the CoL for the quadrotor perching task.}
    \label{fig:airsim_col_diagram_compact}
\end{figure}

\section{Implementation}
The next sections address the experimental methodology used to evaluate the proposed approach and the implementation of the learning algorithm (shown in Figures \ref{fig:airsim_col_diagram_compact} and Algorithm \ref{alg:algo}). 

\subsection{AirSim Environment}
\begin{figure}[!t]
    \centering
    \includegraphics[width=1.0\linewidth]{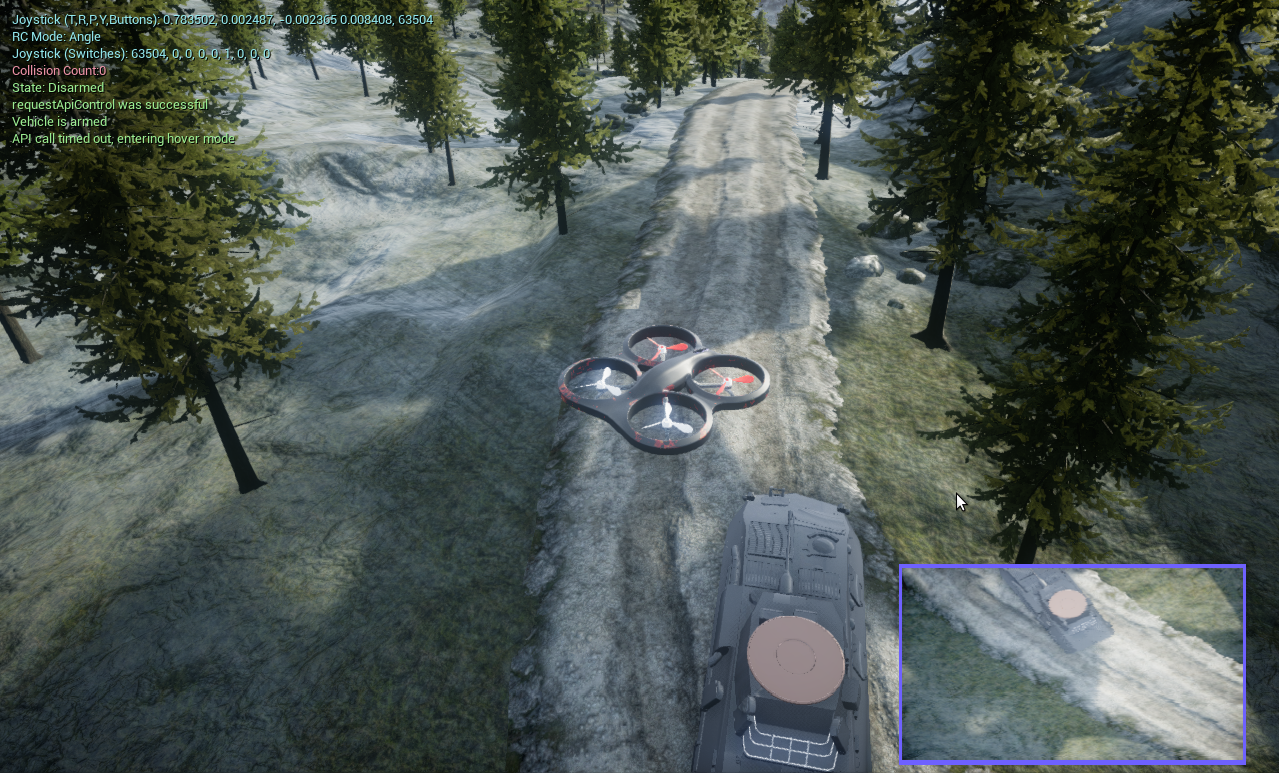}
    \caption{Screenshot of AirSim environment and landing task. Inset image in lower right corner: downward-facing camera view used for extracting the position and radius of the landing pad which is part of the observation-space that the agent learns from. }
    \label{fig:Screenshot}
\end{figure}

We tested our CoL approach (Figure~\ref{fig:airsim_col_diagram_compact}) in an autonomous quadrotor perching task using a high-fidelity drone simulator based on the Unreal Engine called AirSim developed by Microsoft \cite{Airsim2017}. 
AirSim provides realistic emulation of quad-rotor vehicle dynamics while the Unreal Engine allows for the development of photo-realistic environments. 
In this paper, we are concerned with training a quadrotor to autonomously land on a small landing platform placed on top of a ground vehicle (see Figure \ref{fig:Screenshot}). 

The current observation-space consists of vehicle inertial and angular positions, linear and angular velocities, and pixel position and radius of the landing pad extracted using a hand-crafted computer vision module (15 dimensional continuous observation-space).
The vehicle is equipped with a downward-facing RGB camera that captures 320x240 pixel resolution images.  
The camera framerate and agent action frequency is 10.5 Hz, while the human observer views the video stream at approximately 35 Hz.  
The action-space comprises the four continuous joystick commands (roll, pitch, yaw, and throttle), which are translated to reference velocity commands (lateral, longitudinal, heading rate, and heave) by the vehicle autopilot. 

For the perching task, the goal is to land the quadrotor as close to the center of the landing pad as possible. We define a landing a success if the quadrotor lands within 0.5m radius of the center of the platform and a failure otherwise. At the beginning of each episode, the quadrotor starts in a random x,y location at a fixed height above the landing pad and the episode ends when either the quadrotor reaches the ground (or landing pad) or after 500 time-steps have elapsed. 



\subsection{Cycle-of-Learning Algorithm}

\begin{algorithm}[!tb]
\caption{Combining Human Demonstrations and Interventions (Cycle-of-Learning)}\label{alg:algo}
\begin{algorithmic}[1]

\Procedure{Main}{}
    \State Initialize agent's policy $\pi$
    \State Initialize human dataset $\mathcal{D}_H$
    \State Initialize \emph{Update Policy} procedure
    \State Define performance threshold $\alpha$
    \While {task performance $< \alpha$}
        \State Read observation $o$
        \State Sample action $a_{\pi} \sim \pi$
        \If {Human Interaction ($a_H$)}:
            \State Perform human action $a_H$
            \State Add $o$ and $a_H$ to $\mathcal{D}_H$
        \Else
            \State Perform $a_{\pi}$ 
        \EndIf
        \If {End of Episode}
            \State Evaluate task performance
        \EndIf
    \EndWhile
\EndProcedure

\Procedure{Update Policy}{}
    \State Spawn separate thread
    \State Initialize $loss$ threshold $loss_{TH}$
    \While {\emph{Main} procedure running}
        \State Load human dataset $\mathcal{D}_H$
        \If {New Samples}:
            \While {$loss > loss_{TH}$ or $n < n_max$ }
                \State Sample $N$ samples $o, a$ from $\mathcal{D}_H$
                \State Sample $\hat{a} \sim \pi$
                \State Compute $loss =
                \frac{1}{N} \sum_{i}^{N} (\hat{a}_i - a_i)^2$
                \State Perform gradient descent update on $\pi$
            \EndWhile
        \EndIf
    \EndWhile
\EndProcedure

\end{algorithmic}
\end{algorithm}

As shown in Algorithm~\ref{alg:algo} the main procedure starts by initializing the agent's policy $\pi$, the human dataset $\mathcal{D}_H$, the \emph{Update Policy} subroutine, and task performance threshold. 
The main loop consists of either executing actions provided by the agent or actions provided by the human. The agent reads an observation from the environment and an action is sampled based on the current policy. At any moment the human supervisor is able to override the agent's action by holding a joystick trigger. When this trigger is held, the actions performed by the human $a_h$ are sent to the vehicle to be executed and are added to the human dataset $\mathcal{D}_H$ to update the policy according to the \emph{Update Policy} subroutine.

The agent's policy $\pi$ is a fully-connected, three hidden-layer, neural network with 130, 72, and 40 neurons, respectively. The network is randomly initialized with weights sampled from a normal distribution. The policy is optimized by minimizing the mean squared error loss using the Root Mean Square Propagation (RMSProp) optimizer with learning rate of 1e-4.
Unless defined otherwise, the human dataset $\mathcal{D}_H$ is initialized as an empty comma-separated value (CSV) file. Its main goal is to store the observations and actions performed by the human.
The procedure to update the policy in real-time spawns a separate CPU thread to perform policy updates in real-time while the human either demonstrates the task or intervenes. 
This separate thread continuously checks for new demonstration or intervention data based on the size of the human dataset.
If new samples are found, this thread samples a minibatch of 64 samples of observations and actions from the human dataset and is used to perform policy updates based on the mean squared error loss until it reaches the loss threshold of 0.005 or maximum number of epochs (in this case, 2000 epochs).
This iterative update routine continues until the task performance threshold $\alpha$ is achieved, which can vary from task to task depending on the desired performance. For this work, we set $\alpha$ to 1 and only stop training after a pre-specified number of episodes defined in our experimentation methodology to empirically evaluate our method over a controlled number of human interactions, here defined as either human demonstrations or human interventions.

\subsection{Experimental Methodology}
Using the AirSim landing task, we tested our proposed CoL framework against several baseline conditions where we compared against using only a single human interaction modality (i.e. only demonstrations or only interventions) using equal amounts of human interaction time for each condition. By controlling for the human interaction time, we can assess if our method of utilizing multiple forms of human interaction provides an improvement over a single form of interaction given the same amount of human effort. 

Each human participant (n=4) followed the same experimental protocol: given an RGB video stream from the downward-facing camera, the participant controlled the continuous roll, pitch, yaw, and throttle of the vehicle using an Xbox One joystick to perform 4, 8, 12 and 20 complete episodes of the perching task for three experimental conditions: demonstrations only, interventions only, and demonstrations plus interventions with the CoL method, with each condition starting from a randomly initialized policy. For the CoL condition, participants performed an equal number of demonstrations and interventions to match the total number of episodes for that condition. 
For example, given 4 episodes of training, our CoL approach would train with learning from demonstrations in the first 2 episodes and then switch to learning from interventions for the last 2 episodes. 
We compared this to learning from demonstration for all 4 episodes as well as learning from interventions for all 4 episodes. 
This was repeated for 8, 12 and 20 episodes to study the effect of varying amounts of human interaction on task performance. 
Following the diagram in Figure \ref{fig:airsim_col_diagram_compact} and Update Policy procedure on Algorithm \ref{alg:algo}, the agent's policy is trained on a separate thread in real-time, and a model is saved for each complete episode together with the human-observed states and the actions they performed. 
These saved models are later evaluated to assess task performance according to our evaluation procedure described in the next section.

We also compared our approach to a random agent as well as an agent trained using a state-of-the-art reinforcement learning approach. The reinforcement learning agent used a publicly available implementation of Proximal Policy Optimization (PPO) \cite{Schulman2017} with a four degree-of-freedom action space (pitch, roll, throttle, yaw) and was trained for 1000 episodes, using only task completion as a binary sparse reward signal. To investigate the effect of action-space complexity on task performance, we also implemented the PPO where only two actions (pitch and roll) and three actions (pitch, roll and throttle) were available; for both cases, all other actions were held to a constant value. For the two actions condition (pitch and roll), the agent was given constant throttle and descended in altitude at a constant velocity. For both conditions yaw was set to 0. Training time of the reinforcement learning agent was limited to the simulated environment running in real-time. 

\section{Results}


We evaluated our method in terms of task completion percentage, defined as the number of times the drone successfully landed on the landing pad over 100 evaluation runs, for each training method as well as for different amounts of human training data. Additionally, We compared the number of human data samples, i.e. observation-action pairs, used during training for each condition.

\begin{figure}[t]
\centering
  \includegraphics[width=1.0\linewidth]{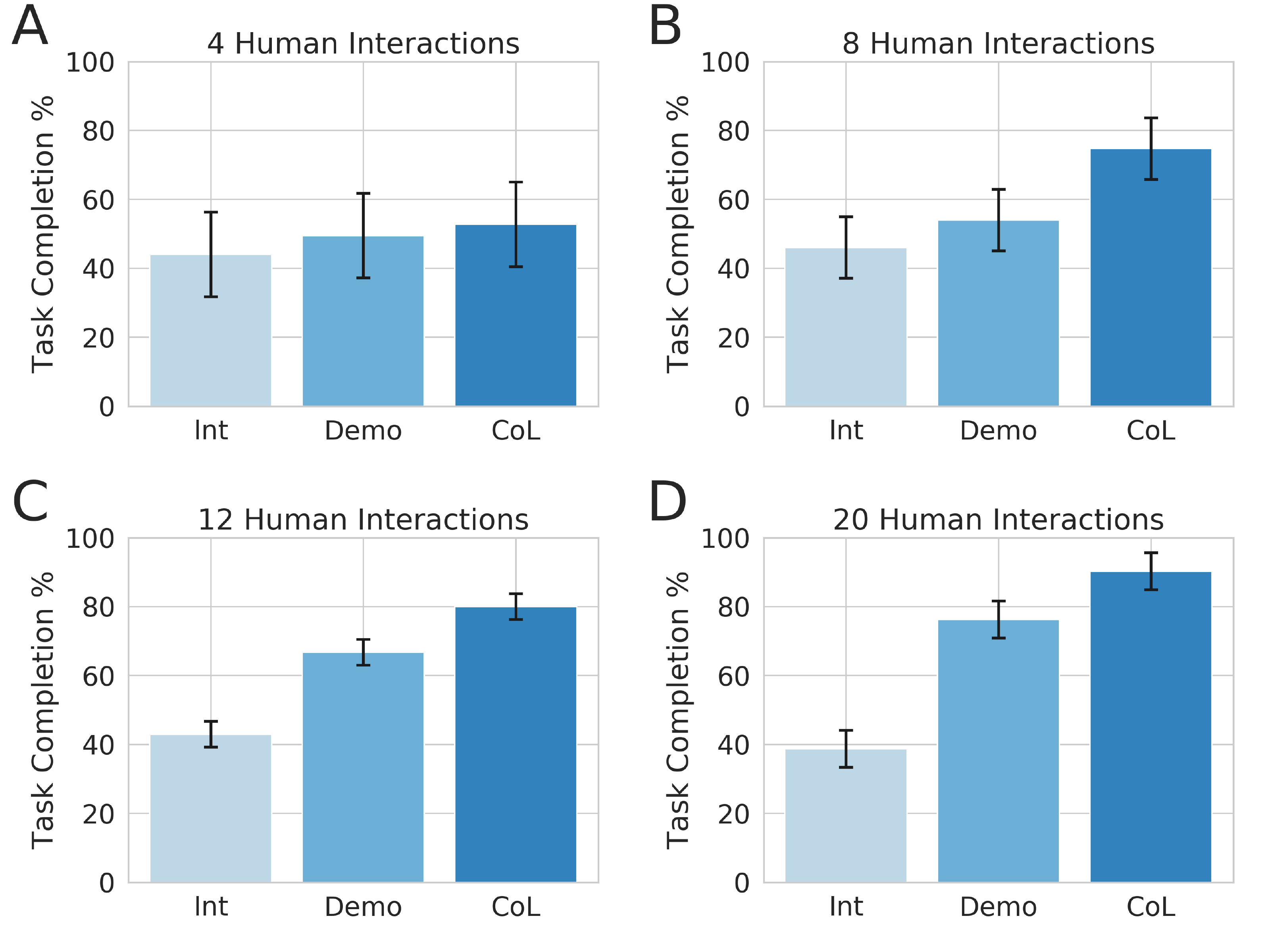}
\caption{Performance comparison in terms of task completion with Interventions (Int), Demonstrations (Demo) and the Cycle-of-Learning (CoL) framework for (A) 4 human interactions, (B) 8 human interactions, (C) 12 human interactions and (D) 20 human interactions, respectively. Here, an interaction equates to a single demonstration or intervention and roughly corresponds to the number of episodes. Error bars denote 1 standard error of the mean.  We see that CoL outperforms Int and Demo across nearly all human interaction levels.}
\label{fig:performance_all}
\end{figure}

Figure \ref{fig:performance_all} compares the performance of the models trained using only interventions (Int), the models trained using only demonstrations (Demo), and the models trained using the Cycle-of-Learning approach (CoL). 
We show results for only these conditions as the random policy condition and the RL condition trained using PPO with the full four degree-of-freedom action space were not successful given the small amount of training episodes, as explained later in this section. 
Barplots show the task completion performance from each condition averaged over all participants with error bars representing 1 standard error. 
Subpanels show the performance for varying amounts of human interaction: 4, 8, 12 and 20 episodes.
For the 4 human interaction condition (Figure \ref{fig:performance_all}A), all methods show similar task completion conditions. 
However, for the 8, 12 and 20 human interaction conditions, we see that the CoL approach achieves higher task completion percentages compared to the demonstration-only and intervention-only conditions, with the intervention condition performing the worst. 

For the final condition of 20 episodes our proposed approach achieves 90.25\% ($\pm$ 5.63\% std. error) task completion as compared to 76.25\% ($\pm$ 2.72\% std. error) task completion using only demonstrations.
In comparison, for the 8 episodes condition, our proposed approach achieves 74.75\% ($\pm$ 9.38\% std. error) task completion in contrast to 54.00\% ($\pm$ 8.95\% std. error) task completion when using only demonstrations. 

\begin{figure}[t]
\centering
  \includegraphics[width=1.0\linewidth]{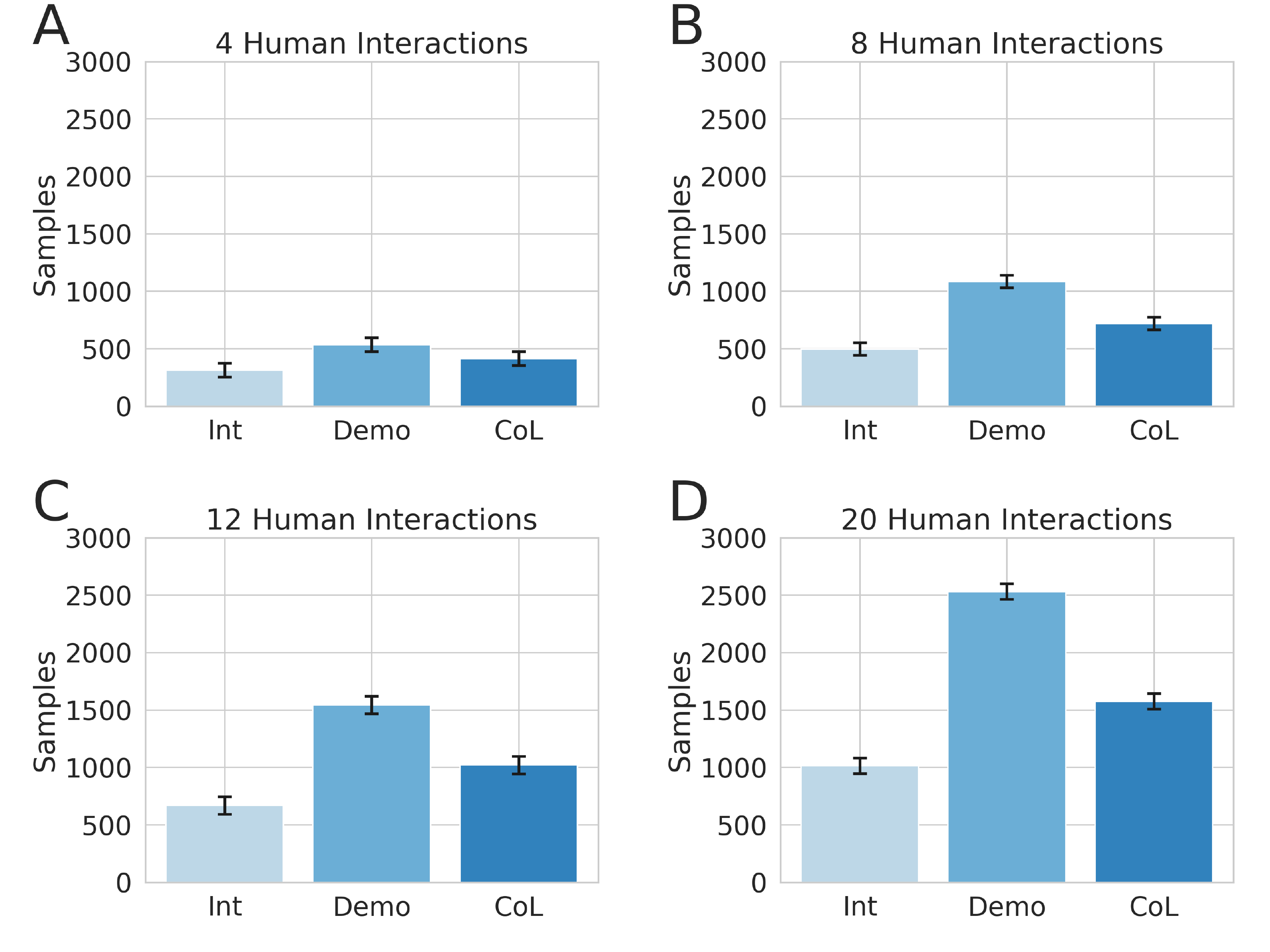}
\caption{Comparison of the number of human samples used for training with Interventions (Int), Demonstrations (Demo) and the Cycle-of-Learning (CoL) framework for (A) 4 human interactions, (B) 8 human interactions, (C) 12 human interactions and (D) 20 human interactions, respectively. Error bars denote 1 standard error of the mean. We see that CoL uses less data than the demonstration-only condition and only slightly more data than the intervention-only condition.}
\label{fig:samples_all}
\end{figure}


Figure \ref{fig:samples_all} compares the number of human data samples used to train the models for the same conditions and datasets as in Figure \ref{fig:performance_all}. 
For the final condition of 20 episodes our proposed approach used on average 1574.50 ($\pm$ 54.22 std. error) human-provided samples, which is 37.79\% fewer data samples when compared to using only demonstrations.
Note that the policies generated from this sparser dataset were able to increase task completion by 14.00\%. 
These results yield a CoL agent that has 1.90 times the rate of task completion performance per sample when compared to learning from demonstrations alone.
This value is computed by comparing the ratios of task completion rate to data samples utilized between the CoL agent and the demonstration-only agent, respectively. 
Averaging the results over all presented conditions and datasets, the task completion increased by 12.81\% ($\pm$ 3.61\% std. error) using 32.05\% ($\pm$ 3.25\% std. error) less human samples, which results in a CoL agent that overall has a task completion rate per sample 1.84 times higher than its counterparts.


\begin{figure}[t]
\centering
  \includegraphics[width=1.0\linewidth]{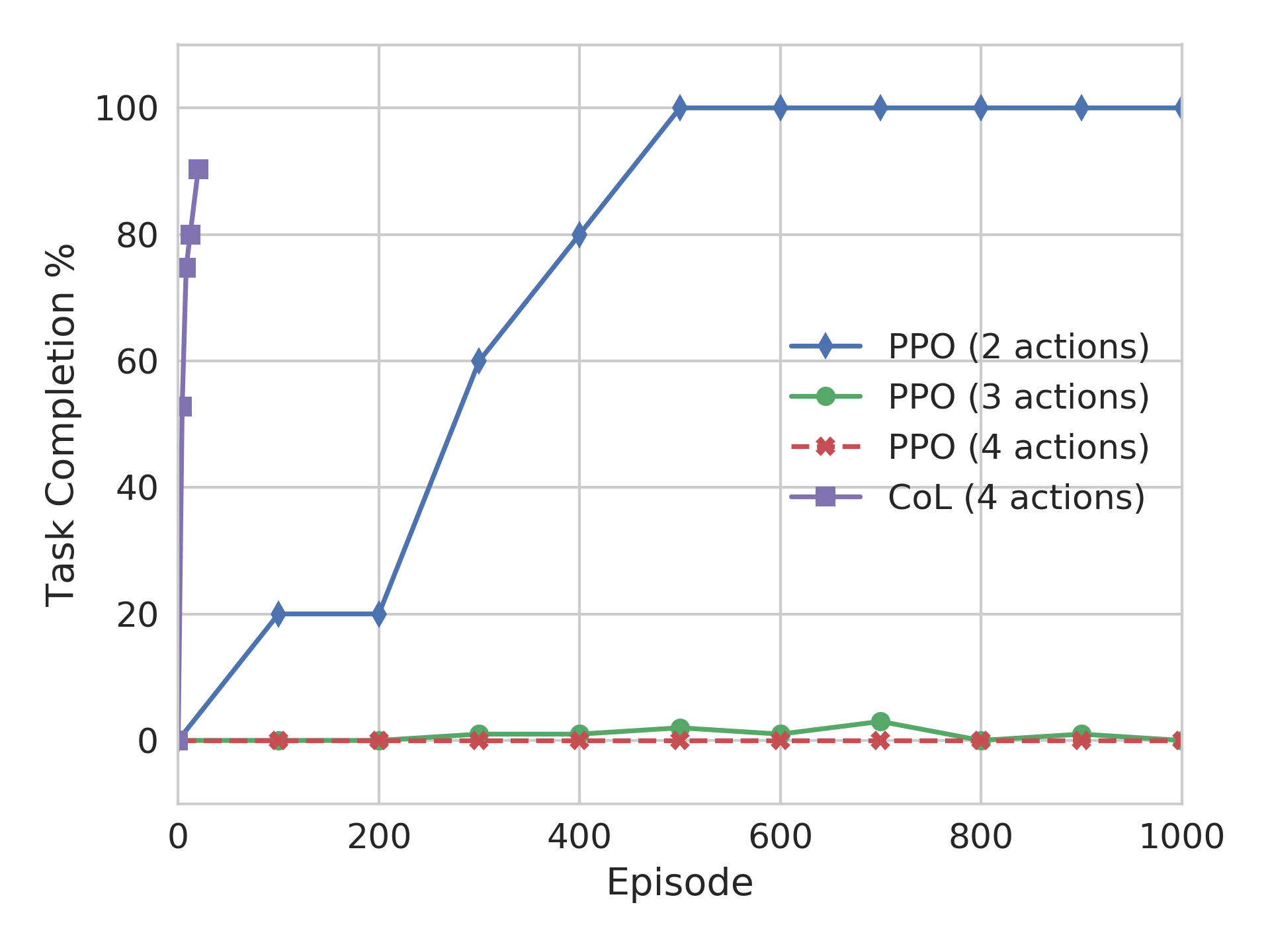}
\caption{Performance comparison between the Cycle-of-Learning (CoL) with four continuous actions and the Deep Reinforcement Learning algorithm Proximal Policy Optimization (PPO) trained for three different task complexities using 2, 3, and 4 continuous actions.}
\label{fig:ppo_curve}
\end{figure}


Figure \ref{fig:ppo_curve} shows a comparison between the performance of the CoL method as well as the PPO baseline comparisons using a 2, 3 and 4 degree-of-freedom action space. For PPO with two actions, the agent was able to achieve 100\% task completion after 500 episodes, on average. However, when the action space complexity increases to three actions, the PPO agent performance was significantly reduced, now completing the task less than 5\% of the time after training for 1000 episodes. As expected, the PPO agent with the full four degree-of-freedom action space fails to complete the task after training for 1000 episodes. In contrast, the CoL method, with the same four degree-of-freedom action space, achieves about 90\% task completion in only 20 episodes, representing significant gains in sample efficiency compared to PPO.

\section{Discussion and Conclusions}

Learning from demonstrations in combination with learning from interventions yields a more proficient policy based on less data, when compared to either approach in isolation.
It is likely that the superior performance of the CoL is due to combining the two methods in sequence so as to leverage their strengths while attenuating their deficiencies.

Having been initialized to a random policy, learning from interventions alone produced more random behaviors, making convergence to a baseline behavior much slower.
Overall performance is thus slower to develop, resulting in lower percent completion for the same number of interaction episodes.
Conversely, learning from demonstrations alone was quicker to converge to stable behavior, but it was consistently outperformed by the CoL across varying numbers of interactions, while having more training data to utilize.
This seems initially counter-intuitive as more training data should result in a more accurate and presumably more proficient policy.
However, as the demonstrations typically follow stable trajectories, the agent is less likely to encounter regions of the state space outside these trajectories.
When enacting the policy at test time, any deviations from these previously observed states is not captured well in the policy, resulting in poor generalization performance.
By allowing the agent to act under its current policy, in conjunction with adaptively updating the policy with corrective human-provided actions needed to recover from potentially catastrophic states, the dataset and subsequent policy is improved.
Thus, the CoL allows for both rapid convergence to baseline stable behavior and then safe exploration of state space to make the policy more refined \emph{and} robust.
    
The results shown in Figure \ref{fig:samples_all} confirm the expectation that the combination of learning from demonstrations and interventions requires less data than the condition of learning from demonstrations alone, for the same number of episodes.
This supports the notion that the CoL is a more data efficient approach to training via human inputs.
When additionally considering the superior performance exhibited in Figure \ref{fig:performance_all}, the data efficiency provided by this technique is even more significant.
This result further supports the notion that a combinatorial learning strategy inherently samples more data rich inputs from the human observer.
 
It should be emphasized here that rather than providing an incremental improvement to a specific demonstration or intervention learning strategy, this work proposes an algorithmically agnostic methodology for combining modes of human-based learning.
The primary assertion of this work is that learning is made more robust, data efficient, and safe through a fluid and complementary cycling of these two modes, and would similarly be improved with the addition of the later stages of the CoL (i.e. learning from human evaluation and reinforcement learning).

As seen in Figure~\ref{fig:ppo_curve}, the PPO baseline comparison method was tested across varying complexity with different numbers of action dimensions. A striking result that can be seen is the significant drop-off in performance when going from two-actions, in which the drone had a constant downward throttle and only controlled roll and pitch, to three and four actions, in which the drone also had to control its own throttle. An obvious characteristic of a successful policy for the perching task is that the drone needs to descend in a stable and smooth manner, which is already provided in the two-action condition, as the downward throttle was set \textit{apriori}. This makes the task of solving for an optimal policy much simpler. In the three and four action condition, however, this behavior must be learned from a sparse reward signal (success or failure to land), which is very difficult given limited episodes. 



When implementing the CoL in real world environments, catastrophic failures may be seriously damaging to the autonomous agent, and thus unacceptable.
Having a human observer capable of intervening provides a mechanism to prevent this inadmissible outcome.
Further, techniques that might be applied to enforce a similar level of safety automatically might limit the exploration of the state space, yielding a less robust or less capable policy.
Analogously to the shift of policy design from roboticists or domain experts to human users and laypersons, which is yielded by human-in-the-loop learning, the technique of learning from interventions shifts the implementation of system fail safes away from developers toward users.
This shift leverages human abilities to predict outcomes, adapt to dynamic circumstances, and synthesize contextual information in decision making.
\subsection{Current Limitations and Future Directions}

Our current implementation is limited to the first two stages of the CoL: learning from demonstrations and interventions.
Our planned future work will include adding in more components of the CoL; for example, learning from human evaluative feedback as done in \cite{Knox2009,MacGlashan2017,Warnell2018}. 
Additionally, we aim to incorporate reinforcement learning techniques to further fine-tune the learning performance after learning from human demonstrations, interventions and evaluations using an actor-critic style RL architecture \cite{Sutton1998}.

A second limitation of the current implementation is that it requires the human to supervise the actions taken by the agent at all times. 
Future work aims to incorporate confidence metrics in our learning models so that the autonomous system can potentially halt its own actions when it determines it has low confidence and query the human directly for feedback in a mixed-initiative style framework \cite{Grollman2007}, similar to active learning techniques. 
Furthermore, our results clearly indicate that a two-stage process - with a primary stage with a large proportion of human-provided actions followed by a secondary stage with a smaller proportion of those actions - outperforms processes with uniformly large or small amounts of human data throughout. 
This suggests there is perhaps an optimal point in the learning process at which to vary in the amount of human input from full demonstrations to interventions.
Figure \ref{fig:performance_all} illustrates this notion across the varying number of interactions shown in the subfigures, i.e. through the change in relative performance between the three conditions.
In future work we will examine if such an optimal mixture or sequencing of demonstrations and interventions exists, such that learning speed and stability are maximized, and if so, whether it is operator dependent.
Rather than having a predetermined staging of the demonstrations and interventions that is potentially suboptimal, a mixed initiative framework could determine this optimal transition point.
This could further reduce the burden on the human observer, allow for faster training, and even provide a mechanism to generate more robust policies through guided exploration of the state space.

This work demonstrates the first two stages of the CoL in a simulation environment with the goal of eventually transitioning to physical systems, such as an sUAS.
The CoL framework was implicitly designed for use in real world systems, where interactions are limited, and catastrophic actions are unacceptable. 
As can be seen in Figure~\ref{fig:ppo_curve}, our method learns to perform the perching task in several orders of magnitude less time than traditional RL approaches, potentially allowing for feasible on-the-fly training of real systems. 
Therefore, we expect that the application of the CoL to sUAS platforms, or other physical systems, should operate in effectively the same manner as demonstrated in this work.
Future efforts will focus on transitioning this framework onto such physical platforms to study its efficacy in real world settings.
One critical hurdle that must be overcome, is the implementation of the learning architecture on embedded hardware, constrained by the limited payload of an sUAS.

Additionally, given that we are utilizing a relatively high fidelity simulation environment, i.e. AirSim, it may be beneficial to bootstrap a real world system with a policy learned in simulation.
Although there are numerous challenges in transferring a policy learned in simulation into the real world, the CoL itself should allow for significantly smoother transfer due to its cyclic nature in which the user can revert to more direct and user intensive inputs at any point during the learning to allow for adaptation to previously unobserved states. 
This capability inherently provides a method of transfer learning in the case of disparities between simulated and real world properties of the vehicle, sensors, and environment.
For example, if the perching behavior learned in simulation was transferred to an actual sUAS, the vehicle dynamics may have unmodeled non-linearities, the imagery may have dynamic range limitations, or the environment may present exogenous gust disturbances.
In such cases, the baseline policy would be monitored and corrected via learning from intervention, if these discrepancies yielded undesirable or possibly catastrophic behaviors.

\section{Acknowledgments}

Research was sponsored by the U.S. Army Research Laboratory and was accomplished under Cooperative Agreement Number W911NF-18-2-0134 . The views and conclusions contained in this document are those of the authors and should not be interpreted as representing the official policies, either expressed or implied, of the Army Research Laboratory or the U.S. Government. The U.S. Government is authorized to reproduce and distribute reprints for Government purposes notwithstanding any copyright notation herein.

\bibliographystyle{aaai}
\bibliography{RL.bib}

\end{document}